%% file: main.tex
\documentclass[prd, aps, 12pt, notitlepage, nofootinbib, floatfix, amssymb, amsfonts, amsmath, eqsecnum]{revtex4-1}

\usepackage[dvipsnames]{xcolor}
\usepackage[pdftex, bookmarks, colorlinks, 
        linkcolor=blue, citecolor=red, menucolor=black, urlcolor=NavyBlue, 
        plainpages=false, pdfpagelabels, hypertexnames=false]{hyperref}
\usepackage{graphicx}
\usepackage{svg}
\usepackage{soul}
\usepackage{chemformula}

\input{definitions.tex}

\begin{document}

\title{Scientific Computing with Large Language Models}
\author{Christopher Culver}
\affiliation{Maxeler Technologies, a Groq Company}
\author{Peter Hicks}
\affiliation{Maxeler Technologies, a Groq Company}
\author{Mihailo Milenkovic}
\affiliation{Maxeler Technologies, a Groq Company}
\author{Sanjif Shanmugavelu}
\affiliation{Maxeler Technologies, a Groq Company}
\author{Tobias Becker}
\affiliation{Maxeler Technologies, a Groq Company}
\maketitle

%\section{Final TODOs}
%\begin{itemize}
%    \item { Cite ANL benchmarking paper }
%    \item { Peter: finish up or leave as is}
%    \item { Decide on format, see below comment }
%\end{itemize}
% can change to two columns by adding the parameter "twocolumn" to the documentclass, font size is also there
%\com{MM wanted two column but then it would have to be 10pt font.  I think this makes the paper look really dense, especially with no figures, tables or graphs.  CC wants to keep it 12 pt one column per page.}

%%%%%%%%%%%%%%%%%%%%%%%%%%%%%%%%%%%%%%%%%%%%%%%%%%%%%%%%%%%%%%%%%%%%%%%%%%%%%%%%%%%
\section{Abstract}
%%%%%%%%%%%%%%%%%%%%%%%%%%%%%%%%%%%%%%%%%%%%%%%%%%%%%%%%%%%%%%%%%%%%%%%%%%%%%%%%%%%
We provide an overview of the emergence of large language models for scientific computing applications.  We highlight use cases that involve natural language processing of scientific documents and specialized languages designed to describe physical systems.  For the former, chatbot style applications appear in medicine, mathematics and physics and can be used iteratively with domain experts for problem solving. We also review specialized languages within molecular biology, the languages of molecules, proteins, and DNA where language models are being used to predict properties and even create novel physical systems at much faster rates than traditional computing methods.

%%%%%%%%%%%%%%%%%%%%%%%%%%%%%%%%%%%%%%%%%%%%%%%%%%%%%%%%%%%%%%%%%%%%%%%%%%%%%%%%%%%
\section{Introduction}
%%%%%%%%%%%%%%%%%%%%%%%%%%%%%%%%%%%%%%%%%%%%%%%%%%%%%%%%%%%%%%%%%%%%%%%%%%%%%%%%%%%
Language is a key component of human communication that has greatly enhanced the evolution of the species.  Using language, humans are able to express ideas, exchange information, and collectively make plans at scales unlike any other in the animal kingdom.  Since the dawn of computing, humans have strived to create AI to automate human tasks.  Recently, large language models~(LLMs) demonstrated the ability to generate text that is often indistinguishable from human-written text. This enables them to aid a wide range of language-oriented tasks such as customer support, document summarization, language translation, coding assistance or content generation.  

As with most AI applications, using LLMs requires two distinct phases: training and inference. Training is the process of feeding large amounts of text data from books, articles, and websites into the model in order to embed syntactical and semantic rules as well as a wide range of knowledge into the model. This process has to occur only once but is computationally demanding, taking weeks to months on high performance computing~(HPC) systems. Inference refers to using a trained model: A query is provided to the model and it will predict an output sequence that will provide a suitable answer to the query. A single inference is less computationally demanding than training, but the challenge lies in that potentially thousands to millions of users interacting with a model will all require inferences to be computed in fractions of a second.
Until recently, LLMs has been hampered by the extreme combinatorial complexity that arises from the large vocabulary of human language and its complex rules. This has made it impossible to generate longer, high quality text sequences.
However, a number of recent innovations have led to a breakthrough in LLM performance. 
Modern LLMs are based on a type of artificial neural network called the transformer~\cite{DBLP:journals/corr/VaswaniSPUJGKP17}, one of the most important innovations in the field of AI in recent years. Central to the transformer architecture is a technique called attention that allows the model to capture long-range dependencies while limiting the combinatorial complexity in longer text sequences. At the same time, models have also increased from millions to billions of parameters, giving them enough capacity to not only understand the syntax, but also the semantics of natural language.
Finally, the introduction of new AI inference accelerators, such as the Groq Language Processing Unit (LPU~\textsuperscript{TM})~\cite{10.1145/3470496.3527405}, has led to significant improvements in throughput and latency during inference. This has made LLMs usable without encountering tedious delays during queries, paving the way for a new design space of real time applications.

Many current LLM use cases focus on direct and simple language-related tasks such as question answering, customer support, chat bots, etc. There is potential for LLMs to be useful in a wider range of application areas including scientific computing. Here, we envision broadly two different approaches: firstly, LLMs could assist scientific processes by being used as a tool on scientific text, e.g., summarization of research papers. Scientific language is noticeably distinct from ordinary language due to using complex noun combinations and a specialized vocabulary.  Within science, different disciplines employ language in different ways, it was shown that biology has imprecise yet densely packed language while physics typically has the opposite~\cite{Persson2016FeaturesAF}.  The description of sciences through natural language is the mechanism the enables humans to collectively solve problems and amass knowledge.  
Secondly, it is also conceivable to treat processes in physics, biology, or chemistry as specialized languages. Molecules, proteins, and DNA all can be seen as languages made up of a specific character set with syntactic rules and semantic meaning completely unlike any ordinary language. It is therefore possible to train special language models just for the purpose of understanding DNA as a language. 
For general languages applied to scientific disciplines and the specialized languages encoding the behavior of physical processes its natural to apply LLMs to these language tasks. 

%%%%%%%%%%%%%%%%%%%%%%%%%%%%%%%%%%%%%%%%%%%%%%%%%%%%%%%%%%%%%%%%%%%%%%%%%%%%%%%%%%%

% \section{Large Language Models}
\section{Sequence modelling architectures}
%%%%%%%%%%%%%%%%%%%%%%%%%%%%%%%%%%%%%%%%%%%%%%%%%%%%%%%%%%%%%%%%%%%%%%%%%%%%%%%%%%%

Originally, sequence modelling tasks were done by RNNs~\cite{rumelhart1986learning}, and LSTMs~\cite{hochreiter1997long}, until the introduction of the Transformer model~\cite{DBLP:journals/corr/VaswaniSPUJGKP17}, which are now ubiquitous for sequence modelling tasks. In this section, we give a brief overview of the historical and current state-of-the-art models in this area. 

The transformer architecture consists of stacked layers of feed-forward networks and attention blocks.  The attention mechanism works as follows: Given the inputs $Q, K, V \in R^{N\times d}$, it computes the outputs $O$ according to $O = \text{softmax}(QK^T)V$, ignoring a normalization factor.  We can interpret the attention mechanism as measuring the similarity between a set of $d$ queries and keys ($d$ is the length of the input sequence), and retrieving a weighted sum of the values corresponding to those keys based on the similarity scores. There are three main variants of the transformer architecture: encoder models, which process input sequences in parallel, decoder models, which generate sequences sequentially, and encoder-decoder models, which take a sequence of inputs and generate a new sequence of outputs.

% Mihailo could list more encoder-only models.
Originally, transformer encoder-only models were popularized by the BERT architecture~\cite{devlin2018bert}.  Encoders can be trained by masked language modelling - removing some words and predicting them based on the surrounding context during training. Other methods such as contrastive objectives like CLIP~\cite{radford2021learning} based models, or classification objectives with sentence transformers~\cite{reimers2019sentence} exist. Encoder style language models have been used for a wide variety of applications, such as clustering
and classification tasks, aligning different modalities in models like LLaVA~\cite{liu2023improved}, Stable Diffusion~\cite{rombach2021high}, and retrieval augmented generation (RAG)~\cite{lewis2020retrieval}.

Another popular framework is auto-regressive language modelling, which generates a sequence $x$ by predicting a sequence of symbols based on the previous symbols in the sequence one at a time, i.e. $p(x)=\prod_{i=1}^{n}p(s_n|s_1,\ldots,s_{n-1})$.
The GPT series of models~\cite{radford2018improving}, which are decoder only transformers, work exactly this way and can be trained on large amounts of unlabelled textual data.

Transformer-based language models have exhibited predictable improvement in perplexity, ability to predict dataset information, both by increasing  the size of the model as well as the amount of data used for training~\cite{kaplan2020scaling}. 
%Performance on downstream tasks can be drastically improved with fine-tuning these models for instruction following~\cite{ouyang2022training}.
This has led to a drastic increase in the capabilities of recent language models, with models spanning trillions of parameters including GPT4~\cite{achiam2023gpt},Gemini~\cite{team2023gemini}, Llama~\cite{touvron2023llama1, touvron2023llama2} and Mixtral~\cite{jiang2024mixtral} achieving state of the art performance in many language modelling tasks.

One of the challenges with pre-trained language models is that the information embedded into the model if fixed and may be lacking specialised domain knowledge. One possible solution is fine-tuning the model: it is a much more lightweight training step where a model can be specialised for certain tasks, significantly improving the performance of the model~\cite{ouyang2022training}. Another important specialisation technique is retrieval augmented generation (RAG) where information from a local database is fed into the model, which enables the model to operate on local information which is not part of the model itself~\cite{lewis2020retrieval}.

A current limitation of transformers is the quadratic time complexity of the attention mechanism, which poses practical limits on sequence lengths and limits a model's ability to understand long distance context.  This is driving research into more efficient operations to replace the attention mechanism, such as the Hyena operator~\cite{poli2023hyena}, sparse matrix multiplications such as MonarchMixer~\cite{fu2024monarch}, and most recently state space models like S4 and Mamba~\cite{gu2023mamba}. Newer architectures achieve very good performance on numerous tasks and even achieve lower perplexity for language modelling compared to transformer models with the same number of parameters.  There are still some tasks such as the copying task~\cite{jelassi2024repeat}, leading to combinations of attention mechanism with more efficient primitives, such as SSM blocks ~\cite{de2024griffin} which were recently scaled up to sizes comparable to leading transformer-based models~\cite{lieber2024jamba}.

Notably, the architectures mentioned above are not limited to only natural language related tasks, but can also model other discrete sequential data of different modalities such as audio~\cite{radford2023robust, lyth2024natural}, images~\cite{peebles2023scalable, dosovitskiy2020image}, video~\cite{liu2024sora}, and the ``language" of biology. %\ref{genomics} or protein sequences \ref{proteins}. 

%%%%%%%%%%%%%%%%%%%%%%%%%%%%%%%%%%%%%%%%%%%%%%%%%%%%%%%%%%%%%%%%%%%%%%%%%%%%%%%%%%%
\section{Molecular Biology}
%%%%%%%%%%%%%%%%%%%%%%%%%%%%%%%%%%%%%%%%%%%%%%%%%%%%%%%%%%%%%%%%%%%%%%%%%%%%%%%%%%%
Molecular biology is the study of living organisms through the interaction of molecules, the building blocks of all materials.  If we can understand how specific molecules dictate the interactions between proteins, then drugs that target specific diseases or viruses can be designed more readily as key ingredients will be identified.  Unfortunately, this link between molecules and their biological utility is so complex that serendipitous drug discovery is still relatively common~\cite{Hargrave-Thomas2012-wt}.  Part of this is due to the intractably complex chemical space of drug like molecules~(about $10^{33}$) that are estimated to be synthesizable~\cite{Polishchuk2013}.  Immense computing and automation efforts are required to explore only a tiny fraction of this domain.  In the following we highlight some applications of LLMs to the language of molecules, proteins, and DNA, and refer to Ref.~\cite{Zhang2024ScientificLL} for a thorough survey.

%%%%%%%%%%%%%%%%%%%%%%%%%%%%%%%%%%%%%%%%%%%%%%%%%%%%%%%%%%%%%%%%%%%%%%%%%%%%%%%%%%%
\subsection{Molecules}
%%%%%%%%%%%%%%%%%%%%%%%%%%%%%%%%%%%%%%%%%%%%%%%%%%%%%%%%%%%%%%%%%%%%%%%%%%%%%%%%%%%
Molecules are a group of atoms held together by chemical bonds, attractive forces between the constituent atoms.  Before trying to physically synthesize a molecule, computations are performed on a candidate molecule to predict its physical properties to ensure it meets the target criteria.  These computations are performed using molecular dynamics~(MD) or density functional theory~(DFT) simulations which require HPC resources.  MD simulations compute the positions of up to billions of atoms while DFT computations simulate only the electrons of the system using a mean field approach.  While there exist efforts to use neural networks to accelerate such algorithms, we focus here on the application of transformers to the language of molecular structure which bypasses these expensive algorithms.  

%\com{CC: Can add more specifics but don't want to get too distracted by HPC.}

The atomic content and even physical structure of a molecule can be represented precisely by a 1-D string, i.e., the chemical formula for water is \ch{H2O}.  A string encoding in principle encompasses all of the properties of the molecule: size, shape, toxicity, 3-D structure, etc since these are the only building blocks for molecules.  Traditionally the properties have to be explicitly computed through the atomic interactions using MD or DFT solvers.  With the recent advancements of transformers to understand not just syntactic but semantic information, its natural to wonder to try and employ transformers to learn the semantic relationship between atomic sequences and physical properties.  The chemical formula above is a bit naive and a more complete molecular description can be specified by, for example, SMILES~\cite{doi:10.1021/ci00057a005} or SELFIES~\cite{Krenn_2020}.  These molecular languages map both the atoms and chemical bonds to characters, for example in SMILES \ch{CO2} is represented as {\thickmuskip=0mu $\text{C(=O)=O}$} where ``$=$" represents a double bond.

The production of enough molecular data to train LLMs has only become available over the last two decades through the use of high-throughput screening~(HTS).  HTS with robotic assistance can currently screen over 100,000 compounds per day producing orders of magnitude more data than previously possible.  There are quite large data sets for both training and bench-marking purposes such as PubChem~\cite{10.1093/nar/gkac956} and MoleculeNet~\cite{Wu2017-xv}.  Much of this data is unlabelled which works well with the self-supervised training language models usually undergo.  These data sets are especially useful for validating a model's ability to predict properties given an atomic string representation.

Encoder style models are primarily used for molecular property prediction and there are many BERT~\cite{Devlin2018} based models due to the large collections of unlabelled chemical data.  SMILES-BERT~\cite{Wang2019SMILESBERTLS} is one such example which is trained using masked language modelling where input molecular strings will be randomly masked.  Applying masking to the pretraining phases enables the model to learn a very general embedding of the molecular space and the role different atoms and bonds play.  After this fine-tuning is applied for classification tasks, SMILES-BERT outperformed other state-of-the-art models in property prediction as of the time of its publication.  One drawback to BERT style models is they focus too heavily on sequence information, causing them to struggle with comprehending molecular structure\cite{Zhang2024ScientificLL}.  Architectures which are also able to learn chemical structure information are emerging through specialised transformers, one based on relative position transformers is MolFormer~\cite{10.1038/s42256-022-00580-7}.  This model was trained on one billion molecules and shown to capture the molecular substructure and spatial interatomic distances.  These kinds of advancements are critical to enabling downstream inference tasks like determining a physiological effect from quantum mechanical properties of molecules.

For designing novel molecules with specific behavior it is common to use GPT like architectures where the transformer will output molecular strings.  A pioneering work was MolGPT~\cite{Bagal2021-rf} which as far as the authors are aware was the first attempt at using GPT architectures on molecular language.  MolGPT is able to be trained on multiple properties and then used to generate novel valid and unique atomic configurations with the desired behavior.  There have since been several advancements based around this architecture, for example cMolGPT~\cite{molecules28114430} which can be used to design molecules that target specific proteins.  By inputting SMILES strings of a target molecule, the generated molecule should interact with, every inference of cMolGPT produces a new molecular sequence, which can be checked against databases for uniqueness.  The cMolGPT model had a 90\% unique compound rate when generating 10,000 valid molecules on three different targets.  These generative models output new molecular compounds but not their properties making it important that encoder only models, like BERT style ones above, continue improving their predictive power.

%%%%%%%%%%%%%%%%%%%%%%%%%%%%%%%%%%%%%%%%%%%%%%%%%%%%%%%%%%%%%%%%%%%%%%%%%%%%%%%%%%%
\subsection{Proteins} \label{proteins}
%%%%%%%%%%%%%%%%%%%%%%%%%%%%%%%%%%%%%%%%%%%%%%%%%%%%%%%%%%%%%%%%%%%%%%%%%%%%%%%%%%%
Proteins are made up of amino acids, organic molecules composed of specific compounds, and perform numerous functions inside living organisms.  While there are hundreds of known amino acids, only 20 are needed to encode the function of proteins and their biological purpose.  Proteins can serve as enzymes, send messages between cells, or provide solid structure in an otherwise fluid environment.  Proteins are created at a molecular level by ribosome which reads the RNA of a cell and produces a 1-dimensional chain of amino acids.  This helps validate the usage of our language of proteins, which are characters representing the amino acids as a 1-D sequence.  Atomic interactions will cause this 1-dimensional chain to \textit{fold} into a 3-D structure after it has been created.  The 3-D structure of a protein is directly responsible for its biological function and is in principle encoded in the textual representation.

% CC: I couldn't find any way to cite this on the Folding@Home webpage...
There are two traditional methods to protein folding, the task of computing its 3-D structure from its textual encoding.  One method is to perform a direct numerical simulation based on the physics of molecular interactions.  Another approach is to use an evolutionary algorithm and do a simulation, which starts with a ``bad" protein and evolves into a useful one.  There are both HPC and distributed computing solutions for these algorithms, one example of the latter is Folding@Home, which anyone can contribute personal computing resources too, and has had tens of thousands of users.

For training LLMs on the language of proteins two resources are UniProt~\cite{10.1093/nar/gkac1052}, which is a hub for protein functional information containing both manually and automatically annotated proteins, and Big Fantastic Database~(BFD) which contains over 2 billion protein sequences.  A general analysis of transformer architectures applied to datasets in UniProt and to BFD was performed by the ProtTrans project~\cite{9477085}.  The project showed that certain architectures were able to perform better than state-of-the-art evolutionary methods while avoiding expensive database searches.  Compared to traditional protein sequence algorithms the LLMs were 5-30 times faster, dependent on model architecture, still the entire human proteome~(20,353 proteins with a median length of 415 amino acids) takes 40 minutes to process. 
%CC: the " human proteom ()" is a direct quote...

The first protein folding LLM to accurately predict atomic resolution structure was AlphaFold~\cite{Yang2023}.  They employed a new architecture using multiple sequence alignment, which highlights homologous features that appear between sequences, which is common in protein strings.  To compute the structure of a protein with 2,500 residues took 18 hours, only after this process can researchers learn about the protein's functionality.  A more recent model is ESM-2~\cite{doi:10.1126/science.ade2574} which uses transformers with up to 15B parameters and avoids multi sequence alignments.  Though there is no substantial improvement to protein structure accuracy there is an almost 60x inference speedup in comparison to other state-of-the-art models. The fast time to 3-D structure prediction will furthermore guide the understanding of how specific proteins have impact at much larger scales~\cite{Tunyasuvunakool2021}.

%%%%%%%%%%%%%%%%%%%%%%%%%%%%%%%%%%%%%%%%%%%%%%%%%%%%%%%%%%%%%%%%%%%%%%%%%%%%%%%%%%%
\subsection{Genomics} 
\label{genomics}
%%%%%%%%%%%%%%%%%%%%%%%%%%%%%%%%%%%%%%%%%%%%%%%%%%%%%%%%%%%%%%%%%%%%%%%%%%%%%%%%%%%

Genomics and transcriptomics are the study of DNA and RNA respectively. These subject areas both aim to deepen our knowledge of the biological macro-molecules they relate to, and then apply this knowledge to various downstream tasks. On a high level these tasks are: the understanding of coding-DNA/RNA and its effects on proteins, and the understanding of non-coding-DNA/RNA and its effects on gene expression and regulation.

% TODO find some numbers and sources
Given the vast scale of DNA and RNA sequences, with sequences of the order 3 billion nucleotides in the case of humans\cite{nrc1988mapping}, the analysis side of genomics and transcriptomics is largely driven by computational algorithms. Classically, these have been deterministic, statistical, and classical machine learning algorithms, however, as the availability of genomics and transcriptomics data \cite{Hood2013} has rapidly grown there has been an uptick in the usage of deep learning models in these areas \cite{Alharbi2022}.  We can consider DNA and RNA to be the languages of life, with their own patterns, grammar and semantic rules. Accordingly, it is no surprise that LLMs, deep learning models typically intended for natural language processing (NLP), have been proven highly effective in the areas of DNA and RNA sequencing analysis.

One of the earlier applications of LLMs to DNA was DNABERT~\cite{Ji2021}, a genomics foundation model able to be finetuned on a variety of downstream tasks.  Unlike with natural language processing where we loosely tokenize on the word or sub-word level, there are many valid tokenization strategies for DNA and RNA.  In the simplest cases tokenization could be done at the single nucleotide level or codon level (3 nucleotide), but more complicated heuristics exist.  The DNABERT authors choose to train multiple models with differing tokenization techniques, splitting on varying length $k$-mers, overlapping sub sequences of a given length. Training for a single model took 25 days with a cluster of 8xNVIDIA 2080Ti GPUs using a masked token replacement strategy.  Finetuned DNABERT models achieved, at the time, state-of-the-art performance in  promoter site, splice site, and transcription binding site prediction which are, broadly speaking, related to gene expression and regulation.

While DNABERT focuses on local aspects of DNA, other works, for example the award winning~\cite{genslm-award} GenSLM~\cite{Zvyagin2022.10.10.511571}, handle entire viral RNA genome sequences.  The GenSLM authors adopt a codon (3 neuclotide bases) level tokenization strategy, with whole viral genomes of the order 30,000 nucleotides, yielding sequence lengths of the order 10,000. GenSLM was pretrained on a dataset of over 110 million prokaryotic gene sequences. This base model was then further trained on whole SARS-CoV-2 genomes\cite{Zvyagin2022.10.10.511571} and adapted for predictive and generative workloads for early warning of variants of concern~(VOCs). In the first case, the model predicts whether a sampled viral genome is likely to be a VOC, i.e., one that is highly aggressive or more harmful. The generative iteration of the model is used to create candidate SARS-CoV-2 genomes to serve as an early warning for potential VOCs~\cite{Zvyagin2022.10.10.511571}. As the size of the sequences being handled by the model and the size of the training data scaled in this project, so did the hardware requirements. 
% Taking out the below, I dont think we should mention training expense since we won't help, and I don't want to mention cerebras for now reason ha
%For model training the GenSLM researches used a combination of traditional and emerging AI supercomputers including Polaris at ALCF, a 560 node system where each node containts 4xA100 GPUs, Selene a similar NVIDIA ran system, and a Cerebras CS-2 system~\cite{Zvyagin2022.10.10.511571}.

Due to the computational challenges that come with the transformer architectures, we are seeing advancements driven by challenges faced in natural language processing trickle down into DNA-LLM research.  One such example of this is in the introduction of hyena layers to genomics LLMs~\cite{nguyen2024hyenadna}. The Hyena layer was introduced to handle long context NLP problems, an issue parallel to the long sequence lengths found when dealing with whole genomes. HyenaDNA~\cite{nguyen2024hyenadna} is trained on sequence lengths of 1 million nucleotides, a scale much greater than the earlier transformer-based DNA-language models. As well as the increased scale, HyenaDNA has achieved state of the art performance on a number of genomics benchmarks.

The focus of DNA/RNA LLMs thus far has primarily been proof of concept, that is to say, although DNA-LLMs are already achieving state of the art performance on genomics-based benchmarks, they are yet to be adopted as common practice in a clinical setting~\cite{Dias2019}. This is largely due to barriers such as model explain-ability and the need for approval by bodies such as the FDA. In this early proof of concept phase the focus of research has been on the rapid training and development of new models, the necessity for large powerful hardware systems to support the training of these models is clear~\cite{Devlin2018,Zvyagin2022.10.10.511571}. However, as DNA-LLM systems move into production the focus will need to shift from fast time-to-product (training) to inference. Accordingly, the hardware requirements for this will need to be addressed, looking toward 
running the DNA/RNA based LLMs on inference focused hardware systems.

% CC opinion : YOu can add a citation if you have one, but given this is more so an academic paper than marketing - we want the readers to draw their own conclusion about which inference provider to use - and we have a citation in the intro about how great we are at LLM inference, we don't need to harp on it
%\com{PH: what citation should we use for Groq here? I'm not sure if this is a bit on the nose. I can't find the sources yet but eventually would be good to include a source about focus on automated and fast systems in NGS projects.}

%%%%%%%%%%%%%%%%%%%%%%%%%%%%%%%%%%%%%%%%%%%%%%%%%%%%%%%%%%%%%%%%%%%%%%%%%%%%%%%%%%%
\subsection{Medical Language Processing} \label{medicine}
%%%%%%%%%%%%%%%%%%%%%%%%%%%%%%%%%%%%%%%%%%%%%%%%%%%%%%%%%%%%%%%%%%%%%%%%%%%%%%%%%%%

% CC: I tried adding a few intro sentences instead, I think this helps give the reader more motivation to know why we mention and cite these things
%\com{PH: I'd maybe like to go into some more detail about some of these.} 

In the medical field there are copious amounts of text written with specialized vocabulary for healthcare workers like doctors and pharmacists.  One of the important features of LLMs is the speed at which they can retrieve and summarize information, which is orders of magnitude faster than a human would require just to gather the appropriate material.
Since 2018 there has been widespread use of BERT and other related LLMs in medical NLP tasks to assist with this problem. Applications of BERT-style transformer models in medicine have included: medical Q\&A bots~\cite{Alzubi2021}, medical text mining/annotation~\cite{Luo2022}, and filtering of public health information~\cite{Nguyen2020}.  

More recently research of generative text-to-text LLMs in the medical domain has become increasingly popular especially due to the relative ease of model fine-tuning.  This has allowed for a high throughput of papers~\cite{Singhal2023,clinicalGPT,ChatDoctor} that investigate the fine-tuning of generative LLMs for medical work. Notably, from Google\&DeepMind fine-tuning their PaLM~\cite{Singhal2023} model achieved state-of-the-art performance on multiple biomedical Q\&A benchmarks. Other works, for example Ref.~\cite{MicrosoftMedPrompt}, have investigated prompt tuning and engineering as methods for enhancing the performance of generative LLMs in the medical domain. There have also been efforts to train generative medical LLMs from scratch on medical papers, abstracts, and clinical guidelines, such as Meditron-70B~\cite{meditron70b}.  Meditron-70B is an open-source model and was able to achieve accuracy within 10\% of the closed-source Med-PaLM-2~\cite{Singhal2023} on medical benchmarks, open-source pretrained models such as this are important as they open up new research avenues, that is to say, anyone can access, fine-tune, and or prompt engineer these models for new medical applications.

%%%%%%%%%%%%%%%%%%%%%%%%%%%%%%%%%%
Another exciting new area of research in medical LLMs is in combined vision language models.~\cite{li2024llava} Vision transformers have already been proven useful in the medical domain, for example, MedVInT~\cite{zhang2023pmc} is a vision transformer that showed good performance on a range of medical classification tasks such as chest x-ray analysis. As such, combined vision-language models similar to the LLaVa~\cite{li2024llava} model could prove an exciting new area of research. Work~\cite{llavamed} is already being done to fine-tune LLaVA models for Q\&A on medical images. In the paper~\cite{llavamed} the authors use the LLM to dicuss a CT scan, the LLM is able accurately describe the location of a lesion in said scan. Due to the visual nature of a large amount of medical diagnosis vision and languagae models are a very appealing area of research.
%%%%%%%%%%%%%%%%%%%%%%%%%%%%%%%%%%

One of the major difficulties in using LLMs in a medical context is the fact that they hallucinate~\cite{Azamfirei2023}. In a setting like medicine, where the stakes are high, clinicians cannot risk an LLM presenting them with incorrect information. Accordingly, methods such as retrieval augmented generation (RAG)~\cite{RAGarticle}, that are known to reduce hallucinations, will be of heightened importance in medical language processing tasks.

% %\com{PH: Todo,  CC: Could leave out?}
% \begin{itemize}
%     \item I need to find some sources but a short para about hallucination and RAG could be good
% %    \item LLMs with visual question answering capabilities for medical images like LLaVA \cite{li2024llava} and MedVInT \cite{zhang2023pmc} 
% \end{itemize}

%%%%%%%%%%%%%%%%%%%%%%%%%%%%%%%%%%%%%%%%%%%%%%%%%%%%%%%%%%%%%%%%%%%%%%%%%%%%%%%%%%%
\section{Math and Physics}
%%%%%%%%%%%%%%%%%%%%%%%%%%%%%%%%%%%%%%%%%%%%%%%%%%%%%%%%%%%%%%%%%%%%%%%%%%%%%%%%%%%
The language of mathematics and physics can be described as ordinary language with mathematical symbols, but this understates the importance of the specialized vocabulary that would not appear in ordinary text.  Mathematical and physics terminology have precise meaning which can be understood with a single sample, which is very different from the usual training of LLMs which learn through many examples.  There are attempts to guide an LLM chatbot to aid researchers through prompt engineering,  one such example is Ref.~\cite{2024arXiv240303154P}.  Here a chatbot is given specialized prompts to guide the LLM towards following the correct analytic steps for applying the Hartree-Fock method, a way to solve quantum many-body problems.  The LLM is able to successfully re-derive 13 of 15 Hartree-Fock Hamiltonians from research papers with only minor mistakes otherwise.  These mistakes can be easily corrected by peer review from the chatbots human counterpart.  Another example of using general LLMs is to guide numerical simulations~\cite{ALEXIADIS2024101721}.  In this work an LLM is prompted to generate a 3-D mesh from a textual description, for example ``Create a bar with a square section centered on the end".  Physical simulations can then be run on the generated meshes, either traditional physics solvers or machine learning powered solutions.
% SS wanted to cite a GNN paper here, but it is a bit out of nowhere...
%, for example Graph Neural Networks (GNN) to simulate dynamics of a mesh~\cite{pfaff2021learning}.

In mathematics there is work to try and solve problems for which brute-force solutions are known, but intractable.  One such is FunSearch~\cite{Romera-Paredes2024}, an LLM that searches for interpretable function formalisms rather than direct solutions to the cap set problem, a combinatorial problem.  The LLM has attached to it a systematic evaluator to search the function space, which is designed to enable a feedback loop with domain experts.  In the cap set problem, FunSearch discovered new constructions of large cap sets, it was applied to online bin packing where it found novel heuristics improving on widely used baselines.  Efforts are also underway to enable LLMs to directly solve mathematical proofs, for example AlphaGeometry proves theorems for Euclidean plane geometry \cite{trinh2024solving}. AlphaGeometry utilizes an LLM to guide a symbolic deduction engine through branching points encountered while writing proofs. AlpaGeometry generates human readable proofs, close to the performance of an average International Mathematical Olympiad (IMO) gold medallist.

\section{Conclusion}
LLMs are being rapidly developed and deployed for software applications that involve natural language processing.  Beyond the world of chatbot style applications there are many scientific domains that employ human language or have their own rigorous scientific language.  Physics, mathematics and medicine use natural language with specialized vocabularies and different semantic meaning from ordinary language and are therefore suitable target for traditional LLMs.  In chemistry and biology, molecules, proteins, and DNA have their own scientific language that represents the underlying physical processes, and they can be targeted with specialised language models.  
%These specialized LLMs need to be trained on domain specific data sets.  

%Inference tasks on these models are expected to have near or above human level capabilities in the natural languages.  
In the domains of mathematics, physics and medicine, LLMs can already be employed to aid researchers by gathering the relevant information and proposing solutions to be reviewed by the human domain experts.  In medicine, this can greatly aid in the ability to sift through the massive amount of textual medical knowledge to focus on only the most relevant information.  In mathematics and physics, LLMs can be used in an iterative process to solve problems too complicated for brute force computation and automate derivations with a well-defined set of steps.  There is also work to train LLMs to perform mathematical proofs, which could enable rapid progression into the understanding of complex mathematical fields, maybe one day generative LLMs will be able to spawn novel mathematical ideas.  

Beyond natural language processing, scientists often encode physical processes in a novel language.  Molecules, proteins, and DNA have special languages which have different syntactic and semantic meaning from natural language.  For molecular and protein research LLMs are able to predict properties from the textual representation as well as design novel structures with target behavior and accomplish this with significant computational speedup. While research using LLMs in such applications is still evolving, it holds promise of enabling entirely new processes of material science, chemistry, and drug design. Finally, in order to serve these wide range of applications, computer systems and services need to evolve to provide token capacity at massive scale, with high throughput, low latency, low energy, and low cost. 
\bibliography{refs}
\end{document}

%% file: definitions.tex
\makeatletter
\renewcommand{\p@subsection}{}
\renewcommand{\p@subsubsection}{}
\makeatother

\numberwithin{equation}{section}

\usepackage{mleftright,xparse}
\NewDocumentCommand\xDeclarePairedDelimiter{mmm}
{%
	\NewDocumentCommand#1{som}{%
		\IfNoValueTF{##2}
		{\IfBooleanTF{##1}{#2##3#3}{\mleft#2##3\mright#3}}
		{\mathopen{\csname##2\endcsname#2}##3\mathclose{\csname##2\endcsname#3}}%
	}%
}
\xDeclarePairedDelimiter{\av}{\langle}{\rangle}
\xDeclarePairedDelimiter{\abs}{\lvert}{\rvert}
\xDeclarePairedDelimiter{\ket}{|}{\rangle}
\xDeclarePairedDelimiter{\bra}{\langle}{|}
\NewDocumentCommand\braket{somm}{%
	\IfNoValueTF{#2}{\mleft\langle #3\,|#4\mright\rangle}{NOTIMPLEMENTED}
}
\NewDocumentCommand\opbraket{sommm}{%
	\IfNoValueTF{#2}
	{\IfBooleanTF{#1}{\langle#3|#4|#5\rangle}{\mleft\langle #3 \left| #4 \right| #5 \mright\rangle}}
	{\mathopen{\csname#2\endcsname\langle}#3\mathopen{\csname#2\endcsname|} #4 \mathclose{\csname#2\endcsname|} #5\mathclose{\csname#2\endcsname\rangle}}
}

\def\beq{\begin{equation}}
\def\eeq{\end{equation}}
\def\beqs#1\eeqs{\beq\begin{split} #1 \end{split}\eeq}